%% file: paper-mo.tex
\documentclass[a4paper]{article}

\usepackage{INTERSPEECH2016}

\usepackage{graphicx}
\usepackage{amssymb,amsmath,bm}
\usepackage{textcomp}
\usepackage{hyperref}
\usepackage{color}

\ninept

\title{Disfluency Detection using a Bidirectional LSTM}

\makeatletter
\def\name#1{\gdef\@name{#1\\}}
\makeatother
\name{{\em Vicky Zayats, Mari Ostendorf and Hannaneh Hajishirzi}}

\address{Electrical Engineering Department\\
	University of Washington \\
  {\small \tt \{vzayats,ostendor,hannaneh\}@uw.edu}
}

\begin{document}

\maketitle
\input{abstract-mo.tex}
\input{intro-mo.tex}

\input{method-mo.tex}
\input{experiments-mo.tex}
\input{conclusion.tex}
\bibliographystyle{IEEEtran}
\bibliography{disfluency}

\end{document}

%% file: abstract-mo.tex
 \begin{abstract}
  We introduce a new approach for disfluency detection using a Bidirectional Long-Short Term Memory neural network (BLSTM). In addition to the word sequence, the model takes as input pattern match features that were developed to reduce sensitivity to vocabuary size in training, which lead to improved performance over the word sequence alone. The BLSTM takes advantage of explicit repair states in addition to the standard reparandum states.  The final output leverages integer linear programming to incorporate constraints of disluency structure. In experiments on the Switchboard corpus, the model achieves state-of-the-art performance for both the standard disfluency detection task and the correction detection task.  Analysis shows that the model has better detection of non-repetition disfluencies, which tend to be much harder to detect. 
\end{abstract}

%% file: intro-mo.tex
\section{Introduction}
A characteristic of spontaneous speech that makes it different from written text -- including informal text -- is the presence of disfluencies. Disfluencies include filled pauses, repetitions, repairs
and false starts. Disfluencies are frequent in all forms of spontaneous speech, whether casual discussions or formal arguments \cite{zayats2014multidomain}.
They present significant challenges for some natural language processing (NLP) tasks on spoken transcripts, such as parsing and machine translation \cite{johnson04,honnibaljoint,mt_disfl}. On the other hand, disfluencies also reflect speaker interaction \cite{shriberg2001errrr}. 
Disfluency detection is most often used as a preprocessing step for NLP, where the goal is removal of the non-fluent word sequences. For extracting information about the interaction, the detection of both disfluent and correction parts can be important. 

A standard annotation of disfluency structure \cite{shriberg94} indicates the reparandum (word or words that the speaker intends to be replaced or ignored), the interruption point (+) marking the end of the reparandum, the associated repair, and an optional interregnum after the interruption point (filled pauses, discourse cue words, etc.)
\begin{quote}
\begin{center}
\fontsize{10}{1}{ [reparandum + $\{$interregnum$\}$ repair] }
\end{center}
\end{quote}
Ignoring the interregnum, disfluencies can be categorized into three types: restarts, repetitions, and corrections, based on whether the repair is empty, the same as the reparandum or different, respectively. Table \ref{tab:Dexamples} gives a few examples.
In this work, we use a slightly modified representation from \cite{ostendorf+13} that distinguishes repetitions (marked by `S') and flattens the nested structure in a sequence of repetitions, which has has led to improved disfluency detection in prior work \cite{ostendorf+13,zayats2014multidomain}.

\begin{table}
\begin{center}
\begin{tabular}{|l|l|}
\hline \bf \quad Type  & \bf Annotation \\ \hline
repair & [ I just + I ] enjoy working \\
repair & [ we + you'd ] have to just \\
repair & [ we want + \{well\} in our area we want ] to \\
repetition & [S it's + \{uh\} it's ] almost like \\
repetition & [S the + th- + the ] decision was \\
restart & [ by + ] it was attached to\\
restart & [ we would like + ] let's go to the \\ \hline
\end{tabular}
\end{center}
\vskip -6pt
\caption{\label{tab:Dexamples} Examples of different types of disfluencies and their annotations. }
\end{table}

Most work on automatic disfluency detection is aimed at cleaning transcripts  for further processing, where only reparandum detection is of interest. In this study we are interested in both the reparandum and repair, motivated by a long term goal of understanding variability in disfluency production related to cognitive load and social context. We introduce a new approach to disfluency detection given text transcripts that leverages a Bidirectional Long-Short Term Memory (BLSTM) neural network and integer linear programming. The model achieves state-of-the-art performance on the standard Switchboard task, and analyses show contributions from including pattern match features in the input.

\section{Related Work}
\label{sec:related}

Approaches to automatic disfluency detection generally fall into two categories: sequence tagging and parsing-based models. Many studies have used a sequence tagging model with begin-inside-outside (BIO) style states that label words as being inside or outside of a reparandum word sequence. The most successful approaches have leveraged discriminative models, including conditional random fields (CRFs) as a classifier \cite{liu06,georgila09,ostendorf+13,zayats2014multidomain,cho2013crf}. In \cite{georgila09}, Integer Linear Programming is integrated with the CRF for optimizing over the prediction sequences. An alternative for improving the CR uses an F-score matching objective, multi-step learner and Max-Margin Markov Networks (M$^3$N) \cite{qian+13}; the objective change had the highest impact. The current best performing system uses a Semi-Markov CRF \cite{semi-markov}.

Another set of approaches leverage parsing and represent a noisy channel relationship between the reparandum and the repair \cite{charniak01,zwarts+10}. The noisy channel parsing models are not well suited to detecting restarts (where there is no repair), but they could be used for identifying the repair, though results on that task have not been reported.  Incremental dependency parsing combined with disfluency removal has also been explored \cite{rasooli2013joint,honnibaljoint}.  Because incremental models do not benefit from reparandum/repair similarity cues, they tend to have lower performance than delayed decision models. Depending on the downstream application, an advantage of parsing models in general is that they jointly optimize for both parsing and disfluency detection performance.  A disadvantage is that they require treebank annotation for training.  Since we are ultimately interested in applying disfluency detection to a broad set of domains, we will leverage a sequence tagging approach, but we extend the label state space to separately model repetitions and repairs, as in \cite{zayats2014multidomain}.


Two recent studies have applied recurrent neural networks (RNNs) to disfluency detection.
One approach explores incremental detection \cite{disfluency_rnn}, with an objective that combines detection performance with minimal latency.
Because of the latency constraints, this approach has weak performance in comparison to other studies on disfluency detection.  Word embeddings learned by an RNN have also been used as features in a CRF classifier \cite{cho2013crf}. In our current study, we also use an RNN, particularly the LSTM framework, but in the standard disfluency detection paradigm (non-incremental), which allows us to use a bidirection architecture and leverage the relatedness of repair and reparandum for repetition and correction disfluencies. Unlike \cite{cho2013crf}, the RNN is the classifier, so our feature embeddings are trained in an end-to-end manner, and they also leverage pattern matching features.

While most studies of disfluency detection focus on using only text transcripts as input, it is well known that prosodic cues are useful in combination with lexical cues \cite{shriberg97,shriberg99,liu06,semi-markov}.
Prosody can carry information that is not represented in transcripts (e.g. length of pauses, fundamental frequency trends), which is relevant for detecting interruption points. However, most studies find that the gain from combining prosodic features with lexical features is relatively small, so our current study focuses on lexical features alone. Adding prosodic information to the existing features is an easy modification with the neural network framework, and we hope to explore this in future work.


%% file: method-mo.tex
\section{General Framework}
\label{sec:methods}


The standard disfluency detection task involving reparandum detection is often called ``edit detection''. The typical sequence tagging model represents 5 states: beginning of the edit region \textit{BE}, inside edit \textit{IE}, the word before the interruption point \textit{IP}, one word edit \textit{BE\_IP} and outside of the edit (including both repairs and fluent regions) \textit{O}. 
For evaluation of edit detection, all words with labels other than \textit{O} are considered edit words.
We also consider two extensions of the state space. The first extension (called explicit repair modeling) includes 8 states, adding: \textit{C} for the repair word, and \textit{C\_IE, C\_IP} for words in nested disfluencies that  belong to both a reparandum and a repair. For the edit detection task, the \textit{C\_IE, C\_IP} states are considered part of an edit region. Note that having explicit repair states does not allow correction detection, as defined in \cite{zayats2014multidomain}, since the repairs associated with repetitions vs.\ corrections are not distinguished. The expanded state space uses the extent of the correction to improve edit detection.
The second extension includes 17 states for joint reparandum and correction detection, expanding all non-\textit{O} states to separately represent repetition and non-repetition disfluencies, as in \cite{zayats2014multidomain}. With this model, we can detect corrections and take advantage of the fact that repetitions tend to benefit from different features than other disfluencies.

As reviewed in section~\ref{sec:related}, CRFs have been used widely for disfluency detection and therefore represent a strong baseline for comparison to the new LSTMs developed here. 
In our work we use the CRF++ toolkit.\footnote{CRF++ available at  \url{https://taku910.github.io/crfpp}}
Starting from a core feature set of lexical, distance-based pattern match features and disfluency language model features used in \cite{zayats2014multidomain} (listed in Table \ref{tab:features}), the CRF features are generated by applying feature functions provided by CRF++ templates to create new features within a relative time frame. For example, using the core feature 'word index,' we can construct n-gram features by applying feature functions across a local time window. A total of 258 features are generated, including combinations of different core features as well as n-grams and POS n-grams.

\begin{table} [t]
\begin{small}
\begin{tabular}{|l|} \hline 
\bf Core Features\\ \hline
1. word index \\ 
2. part of speech (POS) tag \\ 
3. is the word a filled pause \\ 
4. is the word a discourse marker \\
5. is the word a part of an edit word \\
6. is the word incomplete \\
7. distance to the repeated word in the following window \\
8. distance to the repeated bi-gram in the following window \\
9. distance to the repeated word in the preceding window \\
10. distance to the repeated bi-gram in the preceding window \\ 
11. is the POS bi-gram repeated in the following window \\
12. is the POS bi-gram repeated in the preceding window \\
13. is the word and the POS of the next word repeated \\
within the following window \\
14. is the POS and the next word repeated within the \\
following window \\
15. is the word bigram repeated within the next N words \\
allowing some words to come between the two words \\
16. is POS trigram repeated within the N words \\
17. distance to the next used conjunction word \\
18-20. 3 language model features described in \cite{zayats2014multidomain} \\ \hline
\end{tabular}
\end{small}
\caption{\label{tab:features} Core features used to generate CRF features and feature embeddings.}
\end{table}

\section{Proposed Method}
\subsection{RNN Architectures}

\begin{figure}[t]
\includegraphics[scale=0.3]{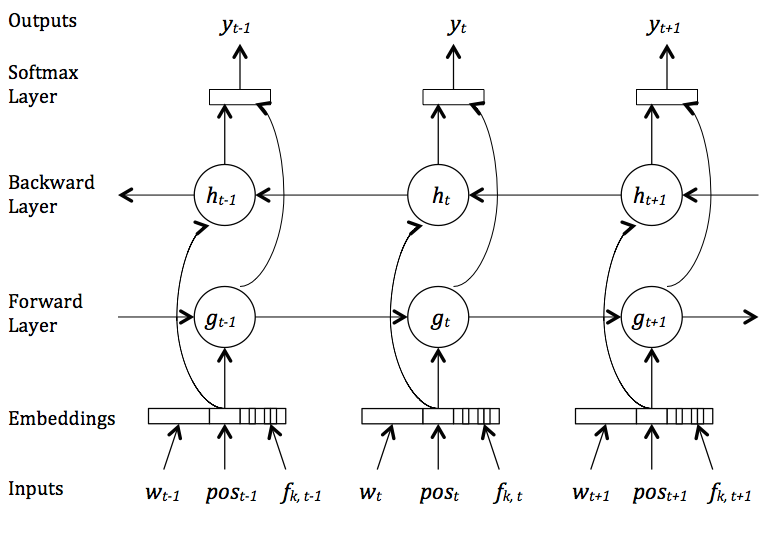}
\centering
\caption{Sample Bidirectional RNN architecture which includes feature embeddings. $w_t$ is a word at time $t$, $pos_t$ is the POS  at time $t$, $f_{k,t}$ is the $k$-th core feature at time $t$ , and $y_t$ is an output state (e.g. \textit{BE\_IP}).
}
\label{fig: birnn_emb}
\end{figure}

We use Long-Short Term Memory (LSTM) RNNs for the task of disfluency detection. LSTMs are deep neural networks to model sequences, achieving good performance in a variety of NLP tasks \cite{mt_lstm,lm_lstm,sutskever2014sequence}. A typical memory cell includes gates to weigh input and history impact at a particular time, allowing the model to determine their relative importance and alleviating the vanishing gradient problem \cite{gradient}. As a result, LSTMs can effectively represent longer phrases, which is useful for the disfluency detection task. For disfluency state sequence tagging, we use a softmax layer at the top layer of the LSTM. 

An LSTM is a directional model and predicts a state given its previous states. For disfluency detection based on text alone, it is difficult to predict a word as disfluent by only observing the previous words prior to the occurrence of the interruption point. Unexpected word sequences following the interruption point and similarity between the repair and the reparandum are important indicators. Therefore, we make use of LSTMs with the input sentence in reverse order. 
%
In addition, we explore use of a bidirectional LSTM (BLSTM) \cite{graves2005framewise}. As shown in Figure \ref{fig: birnn_emb},
the BLSTM uses past and future states in predicting the disfluency tag of a given word. This is particularly useful for predicting both repairs and corrections (8-state and 17-state disfluency). 

\subsection{Feature Embeddings}
 


As shown in Figure~\ref{fig: birnn_emb}, the input vector consists of three main components: word index,  POS tag, and disfluency-based features, as listed in Table \ref{tab:features}. 
The disfluency features provide  useful information about word identity (filler or incomplete words) and patterns (if the exact word appeared previously in a fixed length window). 
%
We separately map one-hot representations of these features to embeddings for a dense representation, and then concatenate them to use as the input to LSTM cells.
For initialization of the word embeddings, we train a backward-LSTM language model on the Switchboard corpus with disfluencies removed. The initialization for POS tag embeddings is similar, with the training text mapped to POS tags. 
All other parameters have random initialization. 
During the training of the whole neural network, embeddings are updated through back propagation similar to all the other parameters.


\subsection{ILP post-processing}
While the hidden states of LSTM and BLSTM are connected through time, the outputs from the softmax layer are not. This often leads to inconsistencies between neighboring labels, sometimes resulting in label sequences that are not valid paths in the state space (i.e.\ `illegal'). 
For example, the model can output the sequence of labels \textit{O} \textit{IE} \textit{IE} \textit{IP}, which is not a valid sequence since disfluencies always start with \textit{BE} tag.  
As such, an additional smoothing over LSTM predictions is needed. Some of the possible approaches include LSTM-CRF \cite{lstm_crf}, Markov model \cite{disfluency_rnn}  or Integer Linear Programming (ILP) \cite{georgila09}. In this work we use the ILP solution previously presented in \cite{georgila09} for disfluency detection. 
Since \cite{georgila09} uses constraints for 5 states only, we collapse the softmax proportions from the larger state space to 5 states, as in
\begin{align}
\begin{split}
 P(y_t=O) &= P(y_t=O) + P(y_t=C)\\
 P(y_t=IE) &= P(y_t=IE) + P(y_t=C\_IE)\\
 P(y_t=IP) &= P(y_t=IP) + P(y_t=C\_IP)
\end{split}
\end{align}
for the 8-state model.

%% file: experiments-mo.tex
\section{Experiments}
\label{sec:experiments}
We assess the proposed BLSTM model for disfluency detection in experiments on the Switchboard corpus of conversational speech \cite{Switchboard}, using the standard division of the disfluency-annotated subset
into training, development and test sets. The flattened version of repetition annotation provided in \cite{ostendorf+13} is used. 
As in other studies on disfluency detection, performance is measured using precision/recall of words in edit regions. In addition, we use the same measure on finer grain labels, including different types of edit regions (repetition vs.\ non-repetition disfluencies) and corrections.

All code is written in theano.\footnote{We modify theano code for LSTM available at \url{https://github.com/JonathanRaiman/theano\_lstm}} 
LSTM (including BLSTM) parameter optimization is done using Adadelta \cite{adadelta} with a mini-batch size of 50. 
We use the Switchboard development set to tune the LSTM parameters (number of dimensions) and to find an optimal stopping point for LSTM training. 
The dimensions of the word embeddings and the hidden dimension are separately tuned for each variation, and the best result is presented in the table. The best number of dimensions for the BLSTM with 17 states is 100, and for all other models it is 150.  The POS embedding dimension is chosen to 5 for all models.
As mentioned previously, word and POS embeddings are initialized using a backwards LSTM language model trained on cleaned-up Switchboard text, and
other model parameters have random initialization. 
All models are trained using only sentences that have 50 or fewer words due to the high computational complexity of longer histories in the LSTM. Fewer than 1\% of Switchboard sentences have length greater than 50 words.
In testing, all sentences are processed.

\subsection{Model Performance}

Table \ref{tab:baseline} shows the CRF and BLSTM performance on the development set for the edit detection task using all three state space alternatives presented in the Section \ref{sec:methods}. 
As shown in the table, the BLSTM has much better performance in edit detection compared to the CRF in all cases.  Moreover, the BLSTM with explicit repair states achieves the best result in the edit detection task, 
which we hypothesize is related to the success of the noisy channel model approach: explicitly representing the extent of the repair allows the model to match the repair to the reparandum for improved detection. 
Table \ref{tab:baseline} also gives the result of the ILP post-processing on the best model. Although some corrections to illegal sequences do not impact edit detection performance, many do. ILP improves the precision of our BLSTM predictions without hurting the recall. 

\begin{table}
\begin{center}
\begin{tabular}{|l|ccc|}
\hline \bf Model  & \bf P & \bf R & \bf F\\ \hline
CRF \textit{5 states}& 91.7 & 78.1 & 84.3 \\ 
CRF \textit{8 states}& 91.3 & 77.6 & 83.9 \\ 
CRF \textit{17 states}& 92.9 & 76.1 & 83.7 \\ \hline
BLSTM \textit{5 states}&  93.6 & 79.0 & 85.7 \\ 
BLSTM \textit{8 states}&  91.5 & 81.5 & \textbf{86.2} \\ 
BLSTM \textit{17 states}&  90.7 & 81.5 & 85.8 \\ \hline
BLSTM \textit{8 states} + ILP&  92.7 & 81.9 & \textbf{87.0} \\ \hline 
\end{tabular}
\caption{\label{tab:baseline} Performance of the LSTM and Bidirectional LSTM on the dev set in the edit detection tasks.}
\end{center}
\end{table}
 
The 17-state models give slightly worse performance for edit detection, but they enable correction detection, the results of which are shown in
Table \ref{tab:correction}. While the BLSTM gives significantly better overall F-score, the two models have very different precision-recall tradeoffs. Augmenting the 17-state BLSTM with ILP post-processing could potentially recover some of the precision lost in moving to the BLSTM.

\begin{table}
\begin{center}
\begin{tabular}{|l|ccc|}
\hline \bf Model  & \bf P & \bf R & \bf F\\ \hline
CRF \textit{17 states}& 73.2 & 37.5 & 49.6 \\ 
BLSTM \textit{17 states}&  57.3 & 51.3 & 54.2  \\ \hline
\end{tabular}
\caption{\label{tab:correction} Correction detection on the dev set using 17 states.}
\end{center}
\end{table}
 
\subsection{Method Comparison}

We evaluate our best models on the test set and compare them to recent methods in the literature leveraging only transcripts. The results on edit detection are shown in Table \ref{tab: test}. For edit detection, we use the explicit repair state space (8 states), which achieves the best results on the development set, including results both with and without ILP post-processing. Both systems beat the best prior result with lexical cues only, achieving state-of-the-art performance of \textbf{$85.9$}.   Looking at the results in terms of the reduction of the performance gap from 1-F of 15.2 to 14.1, this corresponds to a 7\% improvement.
The BLSTMs also beat the higher performing version in \cite{semi-markov} that leverages prosodic features (F=85.4). 
Incorporating prosodic features in a neural network framework is straightforward and will likely lead to an additional gain. 

\begin{table} [t]
\begin{center}
\begin{tabular}{|l|ccc|c|}
\hline \bf Model  & \bf P & \bf 	R & \bf F \\ \hline
Qian et al.\cite{qian+13} & - & - & 84.1 \\ 
Honnibal et al.\cite{honnibaljoint} & - & - & 84.1 \\ 
Ferguson et al.\cite{semi-markov} (lexical)& 90.1 & 80.0 & 84.8 \\ 
BLSTM 8 states& 91.4 & 80.3 & 85.5 \\ 
BLSTM 8 states + ILP& 91.8 & 80.6 & \textbf{85.9} \\ \hline 
\end{tabular}
\end{center}
\caption{\label{tab: test} Comparison of the BLSTM model to state-of-the-art methods in the literature on the test set.}
\end{table}

The 17-state BLSTM model also leads to a significant performance gain in correction detection on the test set, achieving an F-score of 57.7 compared to 49.6 for the CRF result reported in \cite{zayats2014multidomain}, corresponding to a 16\% improvement.

\section{Analysis}

\subsection{Ablation study: CRF vs LSTM}
We also conducted an ablation study to study the effect of the CRF vs.\ LSTM vs.\ BLSTM models in combination with different feature embeddings, with results shown in Table \ref{tab:ablation} for edit detection (5 states) on the development set. 
We compare the  performance of all systems with words alone to the setting when all the features are used. The CRF word features include 1-3 grams within a window of 8 around the word, whereas the LSTM and BLSTM use only the current word index and incorporate longer context through the recurrent structure.
For the \textit{words-only} case, the LSTM and BLSTM give much better results than the CRF. When we add POS and pattern-match features, all systems improve, but the CRF benefits much more. 

\begin{table}
\begin{center}
\begin{tabular}{|l|ccc|}
\hline \bf Model \small{(\textit{input})}  &\bf P & \bf R & \bf F \\ \hline
CRF \small(\textit{words}) & 94.4 & 52.8 & 67.7 \\ 
CRF \small(\textit{words+ pos + feat}) & 91.7 & 78.1 & 84.3 \\ \hline 
LSTM \small(\textit{words}) & 87.6 & 71.4 & 78.7 \\
LSTM \small(\textit{words + pos+ feat}) & 92.4 & 79.0 & 85.2  \\ \hline 
BLSTM \small(\textit{words}) & 87.8 & 71.1 & 78.6 \\
BLSTM \small(\textit{words + pos+ feat}) & 93.6 & 79.0 & 85.7  \\ \hline 
\end{tabular}
\end{center}
\caption{\label{tab:ablation} Comparison of 5-state CRF, LSTM and BLSTM edit detection models with different feature sets  on the dev set. 
}
\end{table}

\subsection{Repetitions vs.\ Non-repetitions}

Repetition disfluencies are much easier to detect than other disfluencies, although not trivial since some repetitions can be fluent. In order to better understand  model performance, we evaluate the 17-state models in terms of their ability to detect repetition vs.\ non-repetition (other) reparanda. The results are shown in Table \ref{tab: repetitions}, showing that the BLSTM is much better in predicting non-repetitions compared to the CRF, allowing better modeling of more complex disfluencies. We conjecture that the dense word representation in the BLSTM captures more of the reparandum/repair ``rough copy'' similarities than the simple POS pattern match features. 

\begin{table}
\begin{center}
\begin{tabular}{|l|c|c|c|}
\hline \bf Model \small{(\textit{input})}  &\bf Repetitions & \bf Other & \bf Either \\ \hline
CRF \cite{zayats2014multidomain} \textit{17 states} & 94.9 & 61.1 & 83.7 \\ 
BLSTM \textit{17 states} & 94.1 & 66.7 & 85.8 \\ \hline 
\end{tabular}
\end{center}
\caption{\label{tab: repetitions} F scores of different types of edits for the CRF and BLSTM on the dev set.}
\end{table}

%% file: conclusion.tex
\section{Conclusion and Future Work}
\label{sec: future_work}

In summary, this paper introduces a Bidirectional Long-Short Term Memory (BLSTM) neural network approach to disfluency detection, 
achieving state-of-the-art performance of \textbf{85.9} F-score on the standard disfluency detection task using explicit repair states, lexical feature embeddings and integer linear programming post-processing. In addition, we improve the state-of-the-art in correction detection.  Analysis shows that performance gain is for cases that are hardest to detect: restarts and repairs. 

The best case BLSTM models leverage hand-crafted pattern match features, indicating that the BLSTM architecture is not sufficiently powerful to learn these cues automatically with the amount of available annotated training data.  An open question for future work is whether other neural network architectures might more effectively learn these cues and thus have a better fit to disfluency detection task. A related question is whether the pattern match features improve performance in cross-domain scenarios, as observed in \cite{zayats2014multidomain}.

